\pdfoutput=1

\documentclass[11pt]{article}

\usepackage{acl}

\usepackage{times}
\usepackage{latexsym}

\usepackage[T1]{fontenc}

\usepackage[utf8]{inputenc}

\usepackage{microtype}

\usepackage{inconsolata}

\usepackage{inconsolata}
\usepackage{graphicx}
\usepackage{subcaption}
\usepackage{algorithm}
\usepackage[noend]{algpseudocode}
\usepackage{float}
\usepackage{optidef}

\title{Benchmark Transparency: Measuring the Impact of Data on Evaluation}

\author{Venelin Kovatchev \\
   School of Computer Science \\
  The University of Birmingham \\
  \texttt{v.o.kovatchev@bham.ac.uk}\\ \\\And
  Matthew Lease \\
  School of Information \\
  The University of Texas at Austin \\
  \texttt{ml@utexas.edu}}

\begin{document}
\maketitle
\begin{abstract}
In this paper we present an exploratory research on quantifying the impact that data distribution has on the performance and evaluation of NLP models. We propose an automated framework that measures the data point distribution across 6 different dimensions: ambiguity, difficulty, discriminability, length, noise, and perplexity.

We use disproportional stratified sampling to measure how much the data distribution affects absolute (Acc/F1) and relative (Rank) model performance. 
We experiment on 2 different datasets (SQUAD and MNLI) and test a total of 135 different models (125 on SQUAD and 10 on MNLI). 
We demonstrate that without explicit control of the data distribution, standard evaluation frameworks are inconsistent and unreliable. We find that the impact of the data is statistically significant and is often larger than the impact of changing the metric. 

In a second set of experiments, we demonstrate that the impact of data on evaluation is not just observable, but also predictable. We propose to use benchmark transparency as a method for comparing datasets and quantifying the similarity between them. We find that the ``dataset similarity vector'' can be used to predict how well a model generalizes out of distribution.

\end{abstract}

\section{Introduction}

 \begin{figure}[h]
     \centering
     \includegraphics[width=\linewidth]{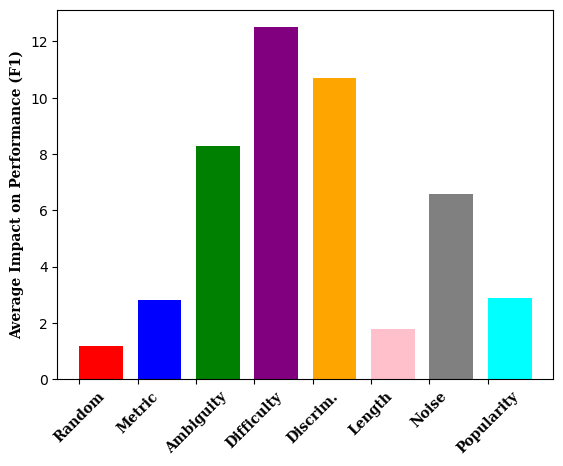}
     \caption{The impact of data distribution on model F1. We report the $\delta$ in F1 caused by re-sampling the test set across each dimension. We report the mean $\delta$ of 125 models on SQUAD. We include random baseline and the impact of changing the ``metric'' from F1 to ``exact''. }
     \label{fig:impact}
 \end{figure}

With the growing popularity and widespread adoption of end-to-end NLP solutions, more emphasis is put on designing and maintaining high-quality evaluation frameworks \cite{NEURIPS2019_4496bf24,srivastava2023imitation,liang2023holistic}. The two key components of model evaluation are \textit{data} and \textit{metrics}. An extensive body of research explores the significance of choosing appropriate metrics \cite{hossin2015review} in various supervised tasks. In this paper, we present \textsc{benchmark transparency}: an automated framework for quantifying the data distribution and measuring the impact data can have on model evaluation.

Figure \ref{fig:impact} illustrates the variance of model performance caused by different data dimensions in the SQUAD dataset \cite{rajpurkar-etal-2016-squad}. To put the results in perspective, we also include the variance caused by uniform random re-sampling and by changing the evaluation metric. It is evident that all data features impact the evaluation more than the random baseline and 4 out of the 6 features are more impactful than changing the metric. 

 \begin{figure*}[h]
     \centering
     \includegraphics[width=\linewidth]{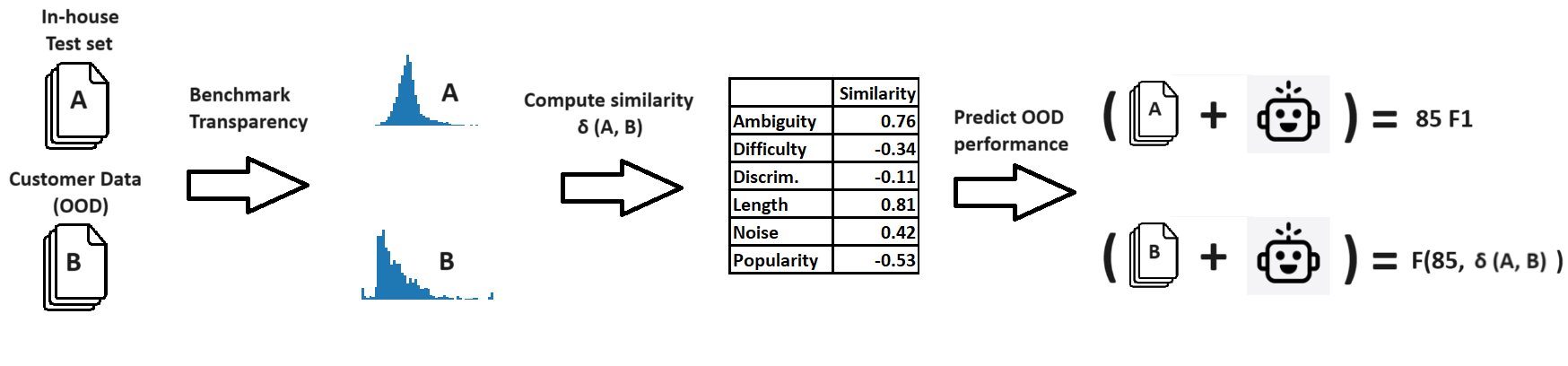}
     \caption{Comparing datasets using benchmark transparency. We measure the data distribution and obtain a ``dataset similarity vector''. The vector can successfully predict the out-of-distribution change of model performance.}
     \label{fig:compare}
 \end{figure*}

A change in F1 by 6 -- 12 points is substantial and statistically significant and puts in question the validity of standard evaluation approaches. We argue that a \textbf{reliable} evaluation framework needs to identify the factors in the environment that largely affect the reported model performance. These factors must be quantified and explicitly incorporated in the evaluation report. We propose \textsc{benchmark transparency} as a way to incorporate scalable data-centric features in model evaluation and subsequently measure and predict the impact of data on reported model performance.

The complexity of linguistic tasks and the importance of data sampling has been discussed before in isolated studies. Lack of sufficient data analysis can lead to discrimination \cite{blodgett-etal-2020-language}, overestimation of model performance on challenging examples \cite{kiela-etal-2021-dynabench,kovatchev-etal-2022-longhorns}, and can hide the errors of the model on particular phenomena \cite{kovatchev-etal-2019-qualitative,hossain-etal-2020-analysis,ribeiro-etal-2020-beyond}. 

We take a more holistic approach, focusing on the overall impact of data on model evaluation. We choose six data dimensions that can be measured automatically: ambiguity, difficulty, discriminability, length, noise, and perplexity. We pose two \textbf{research questions}: 
\textit{1) What is the observable variance in model performance w.r.t. different data dimensions};
and \textit{2) Can data distribution be used to directly compare datasets and predict the variance in model performance}.

We experiment with two datasets: SQUAD and MultiNLI \cite{williams-etal-2018-broad} and evaluate a total of 135 ML models (125 on SQUAD and 10 on MNLI). For the first research question we use disproportional stratified sampling to determine how the absolute (F1/Accuracy) and relative (Ranking) performance of models changes as a function of the data. For the second research question, we split SQUAD and MNLI by domain, using the available meta-data. We then appply \textsc{benchmark transparency} to directly compare the different data splits (Figure \ref{fig:compare}) and use the resulting ``dataset similarity vector'' to predict how model performance will change when applied to unseen out-of-distribution data. We demonstrate that:

\vspace{-4mm}
\begin{itemize}
    \setlength\itemsep{0.1em}
    \item The data distribution has a measurable and statistically significant impact on both absolute (F1/Accuracy) and relative (Ranking) performance of models
    \item The variance in model OOD performance can be predicted if we know the (difference between) source and target data distribution
    \item Our six data dimensions are (empirically) independent. They capture orthogonal aspects of the data and have different impact
    \item There are global tendencies across all models, but there are also significant differences between the individual models 
\end{itemize}

Our findings emphasize the importance of data curation and data sampling in the context of NLP evaluation. Standard evaluation approaches rely on uniform random sampling and make implicit assumptions about the representativeness of the data. We show the impact that these assumptions have on evaluation outcomes, making evaluation inconsistent and opaque. We argue that the assumptions about the data must be made explicit for improved transparency, consistency, and reliability.

\textsc{benchmark transparency} provides clear benefits to various groups of stakeholders and opens promising new lines of research. Incorporating data-centric features can increase the reliability of evaluation, improve the use of NLP benchmarks, and provide a more accurate approximation for OOD model performance. For model developers, the additional feedback on model performance can be used to identify and address model blindspots.

Our approach scales well as it uses simple proxy models to assign data features. It also generalizes to two different NLP tasks: text classification (MNLI) and extractive question answering (SQUAD).

\section{Related Work}

The increased complexity and lack of interpretability of end-to-end neural models makes the design of robust and exhaustive evaluation frameworks a key issue in NLP. Large-scale benchmarks such as  GLUE \cite{wang-etal-2018-glue}, Super-GLUE \cite{NEURIPS2019_4496bf24}, Big-Bench \cite{srivastava2023imitation}, and HELM \cite{liang2023holistic} have been created to address that gap. 

However, existing datasets and evaluation procedures still have issues and limitations. Imbalanced data can lead to issues with bias and fairness \cite{chang-etal-2019-bias,blodgett-etal-2020-language,10.1093/jamia/ocab148}. State-of-the-art models often perform poorly on adversarially generated input \cite{glockner-etal-2018-breaking,wallace-etal-2019-universal,kiela-etal-2021-dynabench}. Some benchmarks can be solved using heuristics and spurious correlations \cite{poliak-etal-2018-hypothesis,mccoy-etal-2019-right}. Models often underperform on linguistic phenomena such as negation \cite{hossain-etal-2020-analysis}, conjunction \cite{saha-etal-2020-conjnli}, or coreference \cite{kovatchev-etal-2022-longhorns}. However standard benchmarks are often ill equipped to capture detailed nuances of model performance \cite{kovatchev-etal-2018-etpc}.
 
Popular benchmarks typically rely on uniform random sampling and statistical aggregation, which can hide model blind-spots on under-represented populations and phenomena. Various strategies have been proposed to improve the evaluation and explicitly identify and address the limitations of the models. \citet{doi:10.1146/annurev-statistics-042720-125902} discuss different metrics that can be used to quantify the bias and fairness of models. The large multi-task benchmarks \cite{NEURIPS2019_4496bf24,srivastava2023imitation} rely on testing a single model across multiple tasks. Datasets designed to test one or more concrete phenomena \cite{kovatchev-etal-2018-etpc,hossain-etal-2020-analysis,saha-etal-2020-conjnli,kovatchev-taule-2022-inferes} can be used for diagnostics, and \citet{ribeiro-etal-2020-beyond} propose an approach for unit-testing NLP models based on predefined capabilities. \citet{kiela-etal-2021-dynabench} suggest the use of ``beat the machine'' human-in-the-loop approach to gather datasets with increasing difficulty \cite{kovatchev-etal-2022-longhorns}. 

More recently, approaches for automatic dataset analysis grow in popularity. \citet{swayamdipta-etal-2020-dataset} analyze the process of model learning and identify patterns in the training set. \citet{rodriguez-etal-2021-evaluation} use evaluation approaches borrowed from the educational domain to improve relative model ranking. \citet{pmlr-v162-ethayarajh22a} propose a measure for dataset ``difficulty'' based on information theory.

While promising, many of the existing approaches for dataset analysis have limited scope and scalability. Some of them are not directly applicable for measuring absolute or relative model performance. We combine and improve existing data-centric techniques and propose new ones with the goal of designing a framework for data-centric and data-informed evaluation of NLP models.


\section{Benchmark Transparency}

\begin{figure*}[htbp]
  \centering

  \begin{subfigure}[b]{0.45\textwidth}
    \centering
    \includegraphics[width=\textwidth]{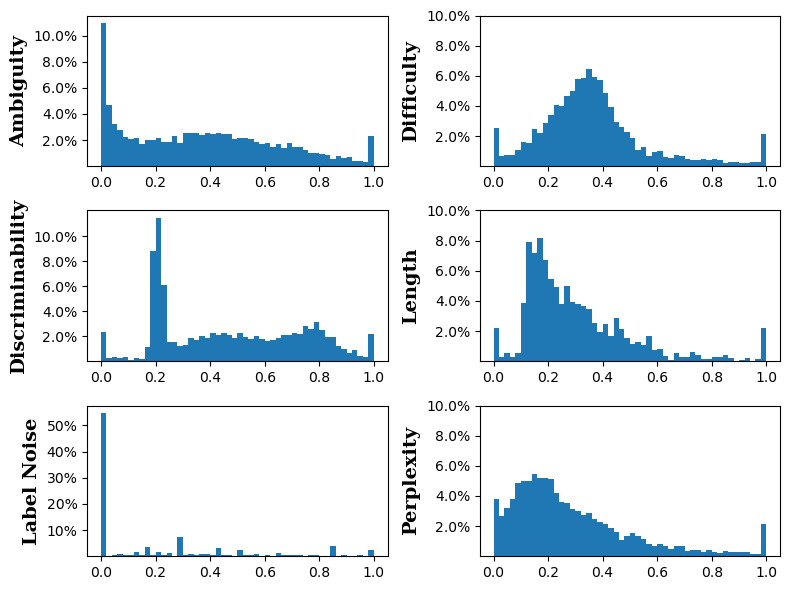}
    \caption{SQUAD}
    \label{fig:data_dist}
  \end{subfigure}
  \hfill
  \begin{subfigure}[b]{0.45\textwidth}
    \centering
    \includegraphics[width=\textwidth]{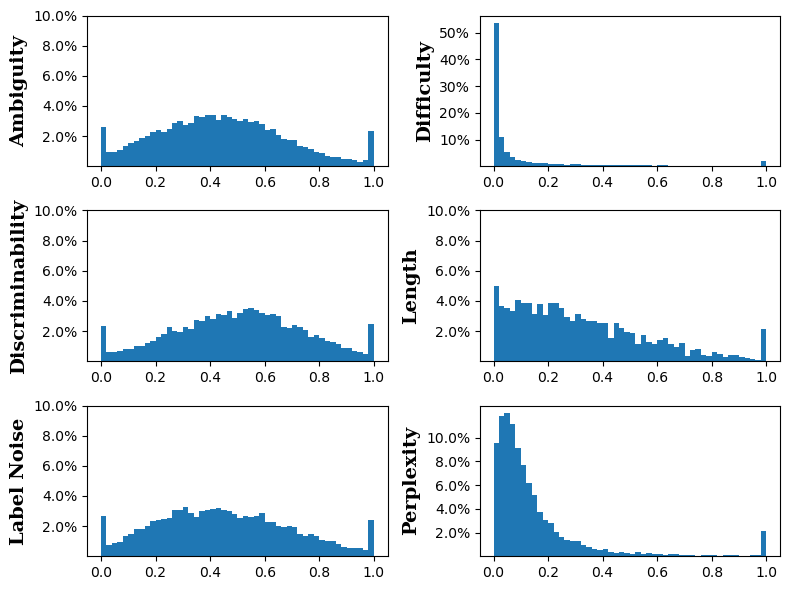}
    \caption{MultiNLI}
    \label{fig:data_corr}
  \end{subfigure}

  \caption{Normalized data distribution of all six dimensions for SQUAD and MultiNLI}
  \label{fig:data}
\end{figure*}

In this paper, we adopt the popular claim that evaluation instances are qualitatively different \cite{rodriguez-etal-2021-evaluation}. For example, some instances are more frequent and popular than others, some are more difficult for humans or models, and some are more useful for distinguishing between strong and weak models. Instances can also come from different domains and can refer to different sub-populations. 

Evaluation frameworks in NLP typically ignore the differences between instances and treat them equally. They rely on Uniform Random Sampling and make the \textbf{implicit assumption} that the resulting dataset is representative for the task. In the cases when differences in the data are made explicit, it is often done across a single axis, such as target demographics \cite{blodgett-etal-2020-language}.

In this paper, we want to quantify the qualitative differences between data instances across multiple (independent) dimensions. We aim to externalize the implicit data assumptions and present an \textbf{explicit analysis} of the data distribution. The objectives of this process that we call \textbf{benchmark transparency} are twofold: 1) to better understand and compare the content of datasets; and 2) to provide a ``dataset representation'' that can be used to objectively measure the impact of data on model evaluation. We propose to use six data dimensions:

\textbf{Ambiguity}
Ambiguous examples are \textit{``instances whose true class probabilities fluctuate frequently during training''} \cite{swayamdipta-etal-2020-dataset}. Note that in our framework ambiguous examples express high variability with respect to the model. They are not necessarily ambiguous to a human. To calculate ambiguity, we adapt the code from \cite{swayamdipta-etal-2020-dataset} for extractive QA and use a BERT-base model to score SQUAD and MNLI.

\textbf{Difficulty}
Intuitively, some instances are more difficult than others and processing them requires different capabilities and world knowledge. To measure instance-level difficulty, we adapt Pointwise \textit{V}-information (PVI) \cite{pmlr-v162-ethayarajh22a} for extractive QA. For each of the two datasets, we train two BERT-large models. The first
model is trained normally, using the full input and the gold label. The second model is trained on the gold labels but without input (MNLI) or with partial input (SQUAD). We obtain PVI by comparing the label probabilities of the two models.

\textbf{Discriminability} 
The concept comes from the domain of education. In Item Response Theory (IRT) \cite{rodriguez-etal-2021-evaluation} ``discriminability'' indicates how useful an instance can be in differentiating between models with varying ability. The underlying idea is that instances where models of different ability disagree are more important for evaluation than instances where the models make the same prediction. For SQUAD, we use the implementation and data of \citet{rodriguez-etal-2021-evaluation}. For MNLI, we use the implementation of \citet{Lalor_2023} and analyze the data ourselves.

\textbf{Length}
We introduce length as a non-trivial baseline to determine the degree to which simple quantifiable dimensions of the data can affect the evaluation outcome. We count the number of tokens in the context (SQUAD) or the sum of tokens in the premise and the hypothesis (MNLI).

\textbf{(Label) Noise}
While ``ambiguity'' measures the inconsistency of model predictions, ``noise'' measures the inconsistency of annotator labels (see \citet{baan-etal-2024-interpreting} for discussion). Both datasets include individual annotator labels. We experiment with using reverse inter-annotator agreement directly or training a model to predict noise.

\textbf{Perplexity}
Perplexity measures the likelihood of a text sequence, given a (neural) language model. Intuitively, some examples are more likely to appear in a context. We link perplexity to the colloquial notions of frequency and popularity, which may be important to various stakeholders. We use a pretrained GPT2-large model to measure the perplexity of a question given a context (SQUAD) and of a hypothesis given a premise (MNLI).

\subsection{Measuring Data Distribution}

\begin{figure*}[htbp]
  \centering

  \begin{subfigure}[b]{0.49\textwidth}
    \centering
    \includegraphics[width=\textwidth]{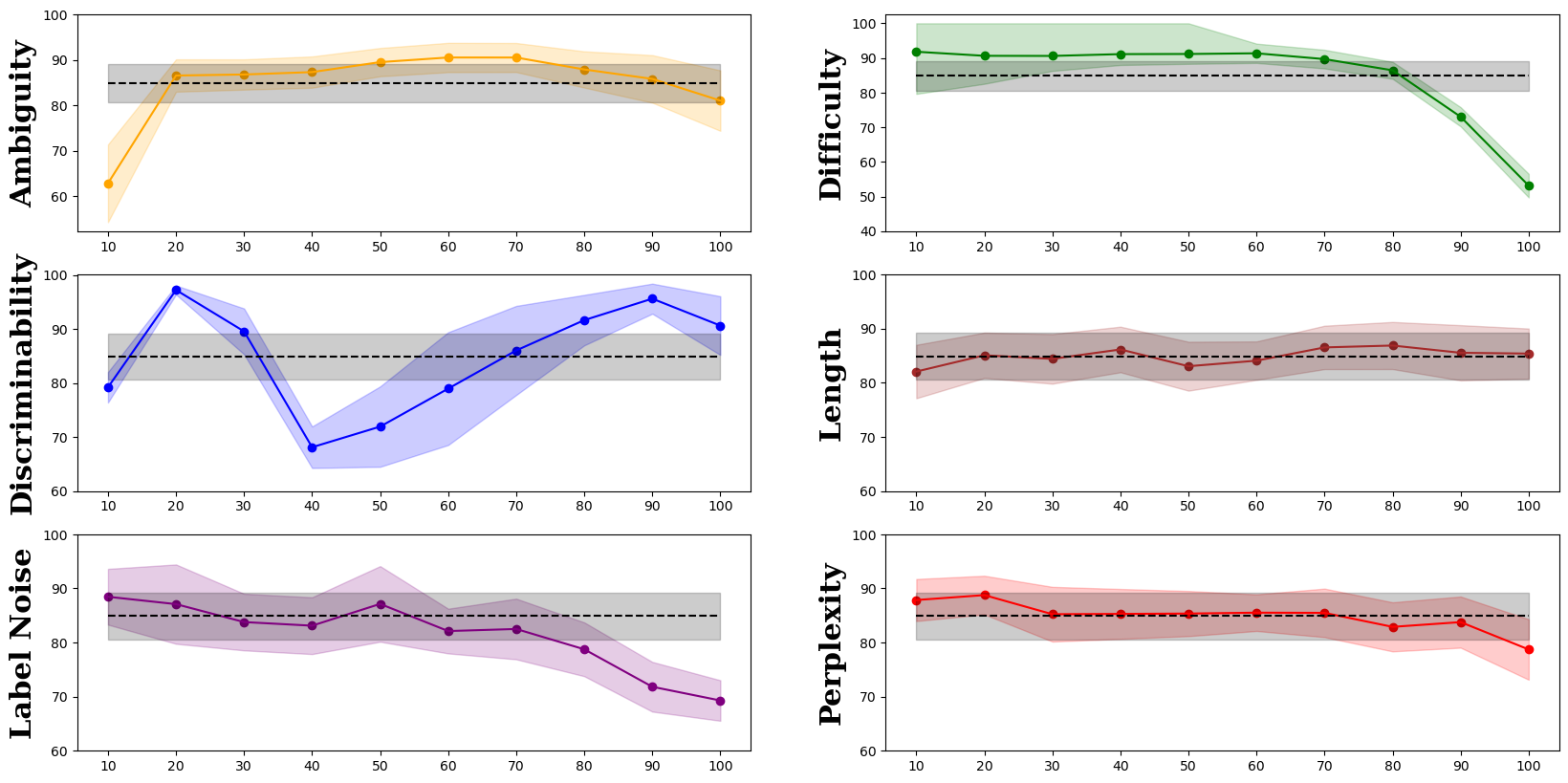}
    \caption{SQUAD}
    \label{fig:data_dist}
  \end{subfigure}
  \hfill
  \begin{subfigure}[b]{0.49\textwidth}
    \centering
    \includegraphics[width=\textwidth]{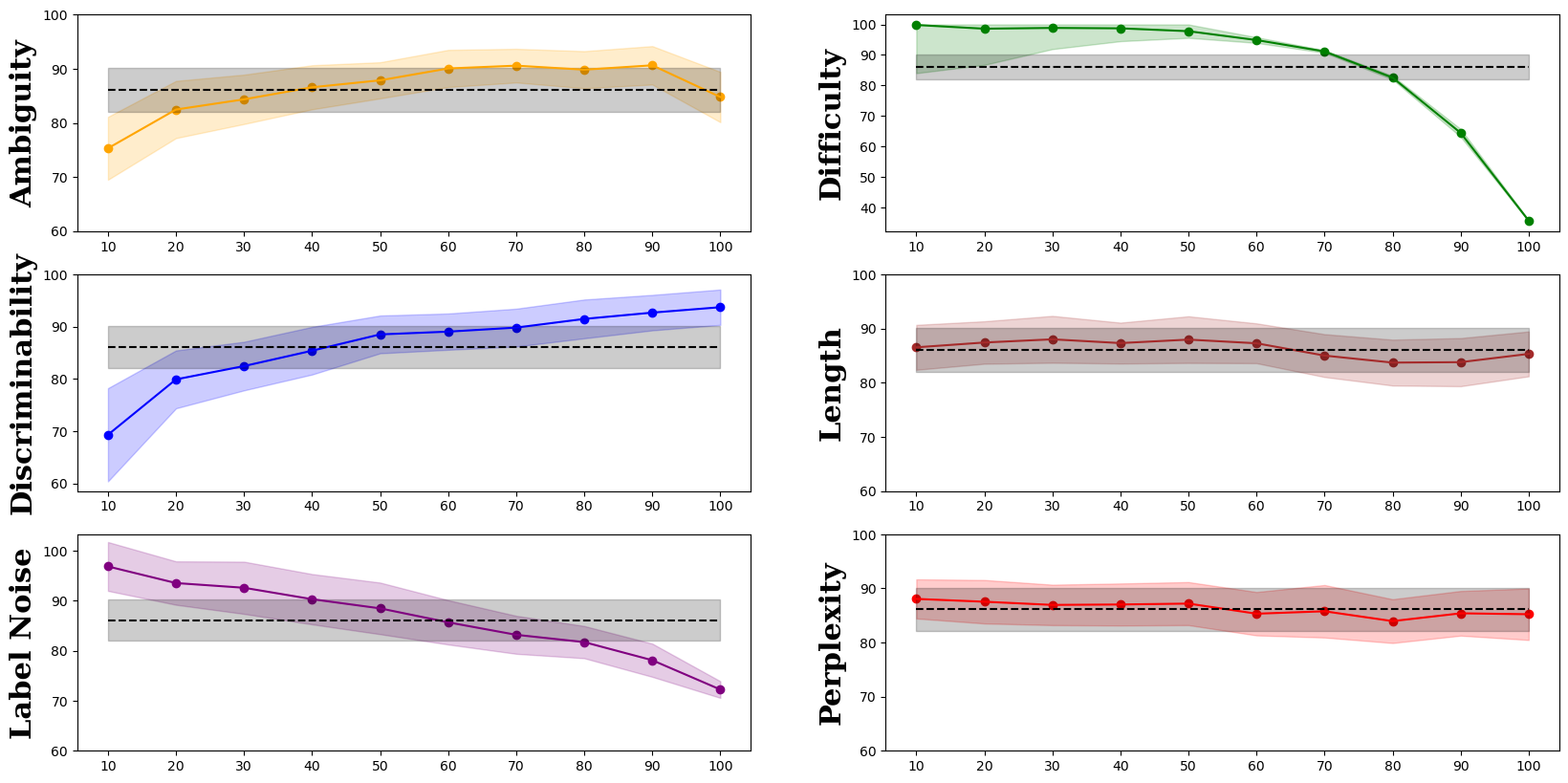}
    \caption{MultiNLI}
    \label{fig:data_corr}
  \end{subfigure}

  \caption{Impact of different data features on model performance (F1) for SQUAD and MultiNLI. On each sub-figure we plot the aggregated change in F1 of all different models (colored shape) as we increase the feature intensity (e.g., as instances become more difficult). The gray region represents the expected random variance at \textit{p \textless{} 0.05}.}
  \label{fig:DIS}
\end{figure*}

A key property of the 6 data dimensions is that they assign a continuous value to every instance in the dataset\footnote{See Appendix \ref{app:features} for formulas and implementation details. The data and code are available at: \url{https://github.com/venelink/benchmark_transparency}}. We can then directly measure the distribution of the features and their inter-correlations. 

Figure \ref{fig:data} visualises the distribution of all features for SQUAD and MNLI. We can observe differences between the individual distributions within each dataset and also between the two datasets. These results indicate that the data profile of the two datasets is different and we cannot draw trivial conclusions. Despite some visual similarities between the distributions, we found no statistical correlation between the features in either dataset. 

Our approach for quantitative data analysis has several practical advantages:

\vspace{-2mm}
\begin{itemize}
    \setlength\itemsep{0.1em}    \item \textbf{scalable}: The process is automated and requires little human supervision. The features are extracted using simple models, relatively small by today's standards. As such the analysis is inexpensive and scales with data size. 
    \item \textbf{task-agnostic}: We apply \textsc{benchmark transparency} to two different supervised tasks: NLI and Extractive QA. The method can be adapted to most supervised tasks. 
    \item \textbf{multi-dimensional}: the features that we use are non-correlated and we argue that they measure different aspects of the data.
\end{itemize}

\section{Data Impact on Evaluation}\label{sec:observe}


The evaluation frameworks of ML and NLP typically report two types of model performance: 1) absolute performance (i.e.: \textit{``How well can we expect a model to  perform on unseen data''}) and 2) relative performance (i.e.: \textit{''How good is each model compared to the alternatives?''}). In this section, we quantify the impact of data distribution on both types of model performance. We use disproportionate stratified data sampling and statistical analysis to address our first research question RQ1: \textit{''What is the observable variance in model performance w.r.t. different data dimensions?''}.

\paragraph{Disproportionate stratified data sampling} For each of the 6 dimensions, we sub-sample the original data to obtain 10 test sets with strictly increasing feature intensity. For example, ``Len\_0'' contains the 10\% shortest examples, and any instance in ``Len\_2'' is longer than any instance in ``Len\_1". We then perform model evaluation on each new test set. As the model parameters and evaluation metrics remain fixed, any difference in model performance can be attributed to the data distribution.

\paragraph{Expected random variance} To put the results in perspective and to calculate the statistical significance of any observed change in reported performance we introduce ``expected random variance'' baseline. We randomly sample 200 test sets with size equal to 10\% of the original data. We test the models on all 200 ``random'' sets and use bootstrapping to determine the two-tailed \textit{p \textless{} 0.05} thresholds for change in absolute or relative performance. The random baseline allows us to filter any fluctuations due to noise or to reducing the test size\footnote{See Appendix \ref{app:boot} for implementation details on stratified sampling and calculating statistical significance.}.

\paragraph{Evaluated models} For SQUAD, there are publicly available instance-level predictions of over 100 different models\footnote{ \url{https://rajpurkar.github.io/SQuAD-explorer/}}. We use that data as-is to calculate absolute and relative model performance without having to re-train the models. We use the data from 125 submissions, with performance between 77 and 92 F1. For MNLI, we implement and evaluate 10 different models\footnote{See Appendix \ref{app:models} for the list of all models and the implementation details (hyperparameters, hardware, and cost).}. 


\subsection{(In-)Consistency of Absolute Performance}\label{exp:absolute}
The absolute model performance is measured with metrics such as Accuracy and F1 and is an approximation of how well the model would generalize to unseen examples. In academic research, the emphasis is often on model ranking, and absolute performance can be overlooked. However in practical applications, the ability to reliably predict how well a model would perform on new data is critical.

In Figure \ref{fig:DIS} we visualize the change in model F1 in the two datasets. The x-axis corresponds to feature intensity: moving from left to right, we plot the performance of models on input with increasing feature intensity (e.g., instances with higher difficulty). The solid line is the mean F1 score of all models and the colored shade around the line corresponds to standard deviation of model score. The gray region represents the expected random variance around the mean at \textit{p \textless{} 0.05}.

Looking at the plots, we can observe that for Ambiguity, Difficulty, Discriminability, and Label Noise, the F1 score of models changes substantially as a function of the  data distribution. This is true for both SQUAD and MNLI. Anecdotally, we can also observe score patterns: the increase of difficulty and label noise leads to a reduced performance, models struggle with instances with low ambiguity and perform well on highly discriminable examples. For length and perplexity, the range of model variance mostly overlaps with the expected random variance. The anectotal analysis is similar for both datasets, although there are discrepancies. For SQUAD, the impact of discriminability does not follow a clear direction and perplexity does not fully overlap with random variance.

\paragraph{Statistical Significance of Performance Variance}

%

The plots in Figure \ref{fig:DIS} indicate that the different data features have a meaningful and substantial impact on model performance, however, to quantify the impact, we perform the statistical tests described in Section \ref{sec:observe} and Appendix \ref{app:boot}. 

\begin{table}[h]
\begin{tabular}{|l||c|c||c|c|}
\hline
Feature & \multicolumn{2}{|c||}{SQUAD} & \multicolumn{2}{|c|}{MNLI} 
    \\ \hline
           & F1 $\sigma$ & 
           \begin{tabular}[c]{@{}c@{}}\% $\delta$ \\\textit{p \textless{} .05}\end{tabular} & F1 $\sigma$ & \begin{tabular}[c]{@{}c@{}}\% $\delta$ \\\textit{p \textless{} .05}\end{tabular}\\ \hline

Ambiguity   & 8.3        & 68\%     & 4.9   & 68\%                                                                             \\ \hline
Difficulty  & 12.5       & 92\%     & 21.0  & 95\%                                                                     \\ \hline
Discr.      & 10.6         & 91\%   & 7.4   & 88\%                                                                    \\ \hline
Length      & 1.8        & 22\%     & 1.8   & 31\%                                                                    \\ \hline
Noise       & 6.6        & 66\%     & 7.7   & 87\%                                                                    \\ \hline
Perplexity  & 2.9        & 33\%     & 1.5   & 17\%                                                                    \\ \hline \hline
Random      & 1.2       & 5\%       & 1.0   & 5\%
                              \\ \hline \hline
Metric      & 2.8      & n/a        & 0.1   & n/a \\ \hline
\end{tabular}
\caption{Impact of data sampling on individual models. We report the standard deviation of F1 w.r.t. different features and the \% of F1 scores that are significantly different from expected random variance.}
\label{tab:DIS-score}
\end{table}

Table \ref{tab:DIS-score} presents the experimental results. 
The $\sigma$ is the aggregated standard deviation of F1 across the 10 tests and indicates the expected magnitude of the impact that each data dimension has on the evaluation. For example, if two datasets $D_A$ and $D_B$ have a significant difference in the distribution w.r.t. data noise, the performance of an NLI model $M_{NLI}$ is expected to change by 7.7 F1.
The statistical significance column reports the percentage of scores (for all models on all data splits) that are significantly different from random fluctuations. 
We can interpret that column as the likelihood that the F1 score of $M_{NLI}$ on $D_A$ and $D_B$ will differ significantly. 
We include two baselines - the impact of random re-sampling and the impact of changing the evaluation metric. For SQUAD we show the difference between using F1 or ``exact'' mataching as a distance metric. For MNLI we compare using Accuracy and F1 as evaluation metric.

The quantitative evaluation confirms the intuition from the visualisation. Difficulty, Discriminability, Ambiguity, and Noise have a significant impact on model performance across both datasets. Distribution shifts w.r.t. Length and Perpexity are less impactful. The overall tendencies are shared among both datasets, but there are also individual differences. Perplexity is much more important for SQUAD, while Noise is as important as Discriminability for MNLI. For the models that we tested on MNLI, we found no difference when changing the metric from Accuracy to micro or macro F1.

Overall, we can conclude that the models are much more sensitive to changes in the data than they are to changes in the metrics. Considering the high \% of statistically significant score changes, we argue that without explicitly considering data features, standard evaluation frameworks are inconsistent and unreliable. A performance variance $\sigma$ of over 6 points questions the validity of the performance report and its ability to correctly predict how well a model would generalize to unseen data.


\subsection{(In-)Consistency of Model Ranking}\label{exp:ranking}

In academic research, ranking is the more popular evaluation criteria, 
as it is directly linked to achieving ``state of the art'' on popular benchmarks. 
To measure the impact of data distribution on relative model performance, we test the consistency of model ranking on different data samples. 
First, we obtain the full model ranking on 200 random sub-samples. 
Then we use bootstrapping and Kendall's Tau to determine the ``expected random variance of ranking'' and the \textit{p \textless{} .05} thresholds for statistical significance. 
Finally, for each data dimension, we obtain 10 different ranking on sub-samples with increasing feature intensity and calculate the portion of the 10 rankings that are significantly different\footnote{See Appendix \ref{app:boot} for the detailed testing procedure.}.

\begin{table}[h]
\begin{center}
    
\begin{tabular}{|l|p{0.1\textwidth}|l|}
\hline
            & \multicolumn{2}{|c|}{Rankings with $\tau$ at \textit{p}\textless{}.05} \\ \hline
& SQUAD \  & MNLI \\ \hline
Ambiguity           & 3/10      & 0/10  \\ \hline
Difficulty          & 7/10      & 7/10   \\ \hline
Discriminability    & 7/10      & 4/10   \\ \hline
Length              & 2/10      & 0/10   \\ \hline
Noise               & 9/10      & 1/10   \\ \hline
Perplexity          & 1/10      & 0/10   \\ \hline \hline
Random              & 0.5/10    & 0.5/10 \\ \hline
\end{tabular}
\caption{Impact of data sampling on model ranking. For each data dimension we report the number of data samples where the overall ranking is significantly different.}
\label{tab:DIS-rank}
\end{center}

\end{table}

Table \ref{tab:DIS-rank} shows the results of the statistical test for ranking. We can observe that the change in ranking does not have a one-to-one correspondence with the change in absolute performance. There are some common trends, such as the importance of Difficulty and Discriminability, but also ranking-specific tendencies. For example, we can see that Noise is the most impactful feature w.r.t. ranking on SQUAD and is much more important than Ambiguity. With respect to F1 score, the impact of Noise and Ambiguity was comparable. We also note large difference between the two datasets and hypothesize that the smaller number of models that we tested on MNLI (10) makes the ranking more robust. 
Overall we find that the data distribution has less impact on ranking than it has on absolute performance. Nevertheless, the results for Difficulty, Discriminability, and Data Noise indicate clear inconsistencies in standard evaluation.

\section{Predicting Changes in Performance}

In Section \ref{sec:observe} we demonstrated the inconsistency of evaluation frameworks caused by changes in the data distribution. 
With the aim of designing reliable evaluation frameworks, we want to go further and use \textsc{benchmark transparency} to predict the changes of model performance as a function of the distribution shift. This will allow NLP practitioners to directly incorporate data features in the benchmark design and in the evaluation metrics. It will also provide a more accurate approximation for model generalizability to unseen data.

\paragraph{Dataset Similarity Vector} To predict the change in model performance across datasets and data samples, we need to be able to quantitatively compare different data distributions. We do that by obtaining a ``dataset similarity vector''. We calculate the Standardized Mean Difference (SMD) across each of the six dimensions. SMD is defined as follows:

\begin{center}
\begin{math}
    SMD = \frac{\bar{x_{1}} - \bar{x_{2}} }
    {\sqrt{(s_{1}^{2}+s_{2}^{2})/2}}
\end{math}
\end{center}

Where $\bar{x_{1}}$ and $\bar{x_{2}}$ are the mean values of the distributions with respect to a particular feature and ${s_{1}}$ and ${s_{2}}$ are the standard deviations of that feature. When comparing two datasets $D_A$ and $D_B$, we obtain the ``dataset similarity vector'' by measuring the data distribution and calculating SMD across all six data dimensions. 

\paragraph{Using SMD to predict change in performance}
Our second research question is RQ2: \textit{``Can data distribution be used to directly compare datasets and predict the variance in model performance?''}. More formally, we want to learn a function $F_{AB}$(Score, Diff) which takes as an input: 1) the performance of model M on dataset $D_A$ (Score); and 2) the difference between datasets $D_A$ and $D_B$ (Diff). The function makes a prediction about the performance of M on dataset $D_B$.


\paragraph{Obtaining different data samples} 

\begin{figure}[h]
     \centering
     \includegraphics[width=\linewidth]{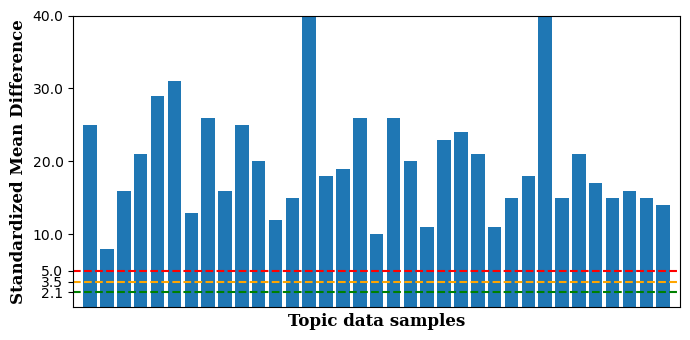}
     \caption{The average SMD between the full SQUAD dataset and different subsets by topic. Dotted lines -- the average SMD between SQUAD and random uniform sub-samples of itself at size 5\%, 10\%, and 20\%.}
     \label{fig:smd}
 \end{figure}

SQUAD and MNLI have explicitly annotated each example with its source domain. We use this information to create different data-samples grouped by data source (henceforth ``topic''). Each of these samples represents a different domain and a different naturally occurring data distribution. Figure \ref{fig:smd} shows the average absolute SMD between each "topic" sub-sample and the full dataset
for SQUAD. 
The three dotted lines shows the average SMD between the full dataset and random uniform samples at size 5\% (5), 10\% (3.5), and 20\% (2.1) and the full dataset. 
It is evident that that \textsc{benchmark transparency} exhibits \textbf{in-distribution consistency} and \textbf{out-of-distribution sensitivity}. This means that we can approximate the full data distribution by using a small sample of in-domain data. At the same time, the data dimensions are able to capture the naturally occurring distribution shifts between independent out-of-domain samples.


\paragraph{Obtaining train and test sets}

We use the following procedure to obtain the data for our experiment:

\begin{enumerate}
    \item Calculate the absolute performance P($M_i$,$D_t$) of all models $M_i$ on all ``topic'' datasets $D_t$
    \item Select source datasets $D_A$ and target datasets $D_B$. For SQUAD, we use the ``full'' dataset as a source and all the ``topic'' datasets as a target. For MNLI, due to the small number of topics (5) and models (10), we make all possible source--target pairings.
    \item Calculate the ``dataset similarity vector'' Sim($D_A$,$D_B$) for every source--target pair
    \item Create individual instances in the format : \\
    <x = (P($M_i$,$D_A$), Sim($D_A$,$D_B$)); \\ y = (P($M_i$,$D_B$)> \end{enumerate}

We then split the data into train and test. We randomly select source--target pairs and all instances associated with that pairings are used for testing. For SQUAD, we select 5 pairings (out of 34), for MNLI we select 1 (out of 5). We re-run the experiments 5 times with different train-test splits to reduce the impact of the sampling strategy. 

\paragraph{Predicting model's OOD performance}

We train a Linear Regression model on our data as it can provide a direct interpretation of the importance of the different dimensions. We evaluate the performance using two different measures: Mean Absolute Distance (MAD) and R2 Score. As a baseline, we predict that the performance of the model will be unaffected, that is P($M_i$,$D_A$) = P($M_i$,$D_B$). The baseline corresponds to the standard random uniform sampling assumption, where we measure the generalizability on an in-domain sub-sample.

\begin{table}[h]
\begin{center}
\begin{tabular}{|l|c|c|}
\hline
            & \multicolumn{2}{|c|}{Mean Absolute Distance} \\ \hline
Model       & SQUAD & MNLI \\ \hline   
Transparency    & 4.1 & 0.9 \\  \hline
Baseline        & 5.9 & 2.1 \\  \hline
            & \multicolumn{2}{|c|}{R2 Score} \\ \hline
Model       & SQUAD & MNLI \\ \hline   
Transparency    & 0.49 & 0.92 \\ \hline
Baseline        & 0.21 & 0.59 \\ \hline
\end{tabular}
\caption{MAD (lower is better) and R2 (higher is better) of using \textsc{benchmark transparency} to predict OOD performance compared to a uniform sampling baseline}
\label{tab:OOD}
\end{center}
\end{table}

Table \ref{tab:OOD} presents the results of the experiment. For both datasets using the ``dataset similarity vector'' reduces the MAD error and increases the R2 score. These results indicate that the information about the data distribution can be used for predicting OOD model performance even with a simple metric such as SMD and a simple model like LR. The OOD prediction works better on MNLI than on SQUAD both in terms of absolute values and in improvement over the baseline.

\begin{table}[h]
\begin{center}
\begin{tabular}{|l|c|c|}
\hline
           & SQUAD & MNLI \\ \hline
Ambiguity  & 0.29    & 0.23                                                          \\ \hline
Difficulty & 0.88    & 1.00                                                         \\ \hline
Discr.     & 0.10    & 0.34                                                          \\ \hline
Length     & 0.05    & 0.06                                                          \\ \hline
Noise      & 1.00    & 0.16                                                          \\ \hline
Perplexity & 0.29    & 0.18                                                          \\ \hline
\end{tabular}
\caption{Feature importance in OOD prediction}
\label{tab:feat-imp}
\end{center}
\end{table}

Table \ref{tab:feat-imp} shows the importance of the individual dimensions when predicting the change in model performance. These are the weights of the Linear Regression after applying a standard scaler to the SMD across each dimension and then dividing the weights by the maximum value for visualisation purposes. 
Similar to the observations we made in Section \ref{exp:absolute}, the most important data feature is Difficulty. Noise and Ambiguity are also important for both datasets. Length is of little importance and Discriminability and Perplexity are only impactful for one of the datasets.





The experiments in this section further validate our choice of data dimensions and indicates that \textsc{benchmark transparency} can be used to improve the reliability of evaluation. The data distribution within the same data sample is stable, and when the data distribution shifts in an OOD setting, we can use the dataset similarity vector to anticipate the change in absolute model performance.

\section{Discussion and Conclusions}

In this paper we emphasize and quantify the importance of data in NLP evaluation. 
There are various popular ways of calculating model performance: Precision, Recall, F1, Accuraccy, and AUROC for absolute performance; global ranking, pairwise ``duels'' \cite{liang2023holistic}, or complex statistical models  \cite{rodriguez-etal-2021-evaluation} for ranking.
We argue that the specifics of the test data are no less important than the choice of adequate distance and aggregation metrics. The effect that data has on model performance is, no doubt, known intuitively by most researchers. However, to the best of our knowledge, this is the first systematic and multi-dimensional approach towards quantifying data distribution and measuring its impact on evaluation across multiple tasks and models.

\paragraph{Benchmark transparency} We proposed a framework for quantifying and comparing the data distribution of datasets for supervised NLP problems. 
We applied our framework to two different datasets, designed for different tasks: SQUAD and MultiNLI. 
We observed that the difference in data distribution significantly affect both absolute and relative model performance. Our findings are consistent across both datasets and multiple models. We concluded that \textbf{the observed variance is a property of the data}, rather than of the models. 
We further demonstrated that \textsc{benchmark transparency} is not just a tool for data analysis, but can be used to successfully predict the changes in model performance out-of-distribution. 

\paragraph{Choice and importance of data features}

In this paper we proposed six different data dimensions: ambiguity, difficulty, discriminability, length, noise, and perplexity. Our objective was to provide a framework for automated and scalable quantification of data distribution across multiple dimensions. Our experiments indicated that the metrics are empirically independent and impact the model performance in a different way. The data features can be extracted at a low computational cost as they typically require simple proxy models. The data distribution is relatively consistent within the dataset, which means that it can be approximated by sampling only a portion of the data. Our choice of data features was empirical rather than theoretical and is non-exhaustive. We encourage the community to experiment with more data features and with alternative ways for calculating the existing ones.

\paragraph{Reliable evaluation for NLP}

A high quality evaluation frameworks need to be reliable. They need to consider and control all factors that significantly and systematically impact the evaluation outcome. Testing and reporting complementary results using different metrics is a standard practice in NLP. However data centric approaches to evaluation are less popular. We have demonstrated that data distribution is key factor in model performance and via \textsc{benchmark transparency}, we have provided the community with a tool for quantifying, conditioning on, and controlling the data distribution\footnote{All of our code and data are available at \url{https://github.com/venelink/benchmark_transparency}}.

\paragraph{Error analysis and model improvement}

 \begin{figure}[h]
     \centering
     \includegraphics[width=\linewidth]{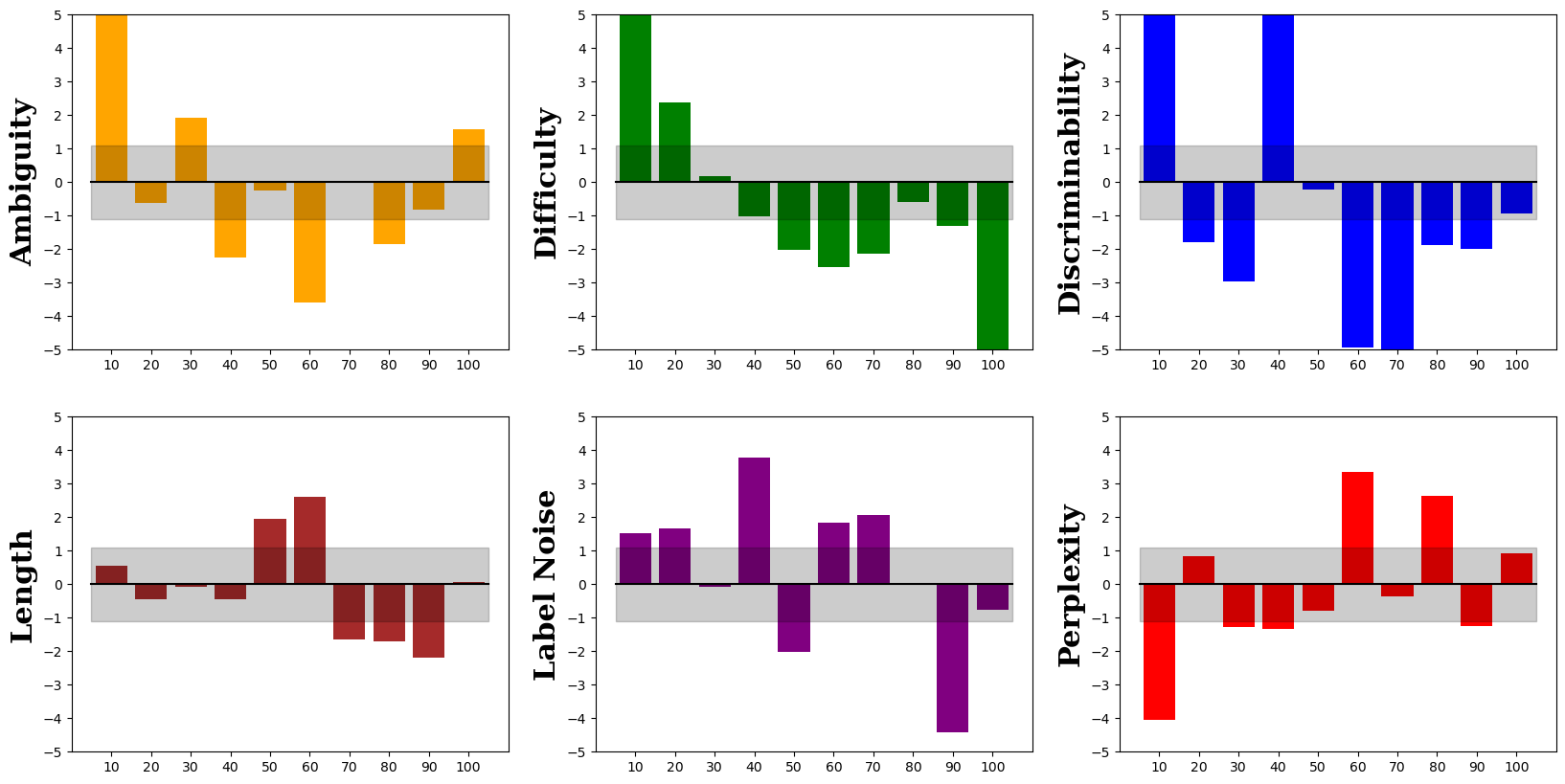}
     \caption{Comparison of two models with identical F1 on the full SQUAD dataset. Each sub-figure shows the difference (in F1) between the two models on datasets with varying feature intensity.}
     \label{fig:single}
 \end{figure}

A data-informed evaluation can also benefit model developers by providing a detailed performance profile with strengths and potential blindspots of the models. Figure \ref{fig:single} compares two of the best performing models on SQUAD. On the full dataset, the two models achieve the same score at 90 F1. We used \textsc{benchmark transparency} and evaluated the two models on data splits with increasing feature intensity as described in Section \ref{sec:observe}. We then calculated the difference in F1 between the models on each split. We can see that despite having identical performance on the full dataset, the models make qualitatively different predictions and have different performance profile. Anecdotally, one of the models seems to excel at easy examples, while the other performs better on hard ones. This information can be important when determining which model to deploy in production or where to focus on model improvement.

\paragraph{Future work}
As a future work, we plan to use the data dimensions to design dynamic benchmarks that can be adapted to stakeholder needs and select examples dynamically based on model performance. We are also exploring the possibility of using the data distribution to guide model training and the development of data-centric loss functions and optimization strategies.

\section*{Acknowledgements}
We thank the reviewers for their valuable feedback, the online workers who provided annotations for data used, and the 
University of Birmingham Tier 2 HPC for their computational resources. This research was supported in part by Good Systems\footnote{\url{http://goodsystems.utexas.edu/}}, a UT Austin Grand Challenge to develop responsible AI technologies. The statements made herein are solely the opinions of the authors and do not reflect the views of the sponsoring agencies.

\section{Limitations}

Our approach and data dimensions are task- and language-agnostic. However, the formal definitions of each each data dimension can be task specific and may be non-trivial. For many of the dimensions, we had to perform an adaptation of formal definitions designed for classification to Extractive QA. Our choice of how to define different dimensions is one of many possibilities and is based on empirical evidences and discussions between the authors. Alternative definitions of dimensions (e.g., difficulty or popularity) may yield different results. 

The data dimensions that we use are designed for scalability and use basic transformer models (BERT, GPT2) to reduce the training time and cost. Prior work has shown that features extracted using BERT correlate strongly with features extracted using state-of-the-art models. Our experimental results confirm the applicability of basic transformer models for the purpose of data analysis. Nevertheless, since the data is based off a single model, it may contain model-specific biases. For a practical implementation, we suggest aggregating the score from two or more models. Furthermore, for particular domains, it may be better to pick a domain specific implementation of a model (e.g. GPT trained on biomedical text). We keep our implementation general.

The data dimension of ``discriminability'' is the only one that does not scale very well with size, as it requires multiple models being trained and tested on the same data. It can be calculated for popular, publicly available benchmarks such as SQUAD, but is use on less popular and/or private datasets may be more computationally expensive.

\bibliography{anthology,custom}

\appendix

\section{Obtaining Data Features}
\label{app:features}

This appendix presents detailed information on the implementation of the different data-centric features and the decision process behind them.

\paragraph{Ambiguity}
For ambiguity, we adopt the definition from \citet{swayamdipta-etal-2020-dataset}: ``instances whose true class probabilities fluctuate frequently during training (high variability), and are hence ambiguous for the model''. To obtain the variability, we:

\begin{enumerate}
    \item Finetune a BERT-large model for 10 epochs. At every epoch we predict the instances in the validation set, keeping the class probabilities  
    \item Take the probabilities of the correct answer at each epoch and store them in a vector called ``conf''
    \item For each val instance we calculate the variability following the original implementation: \\ np.sqrt ( \\ np.var(conf) +  \\np.var(conf) * np.var(conf) / (len(conf)-1) \\ ) \\
    
\end{enumerate}

The original implementation is only for text classification, but we extend it to Extractive QA with reasonable adjustments.

For text classification (MNLI) we use the class probabilities as they are generated by the softmax at the last classification layer. 

For extractive question answering (SQUAD), we obtain ``class probabilities of the correct answer'' by multiplying the probability of the correct start token by the probability of the correct end token and normalize by the probability of all valid start/end pairs. When applied to extractive QA in the format of SQUADv2, we also account for the probability of ``no answer''. 

Note that the original implementation of benchmark transparensy and ambiguity is focused on ``training dynamics'', so the algorithm is designed to score training data rather than test data. However, we extend the concept to scoring validation data at each epoch (for both MNLI and SQUAD, our data analysis is performed on the publicly available validation data).

\paragraph{Difficulty}
For difficulty, we follow the implementation by \citet{pmlr-v162-ethayarajh22a}. To obtain PVI:

\begin{enumerate}
    \item Finetune a BERT-large model for 3 epochs on the train set
    \item Finetune a BERT-large model with the same hyperparameters as in 1), but the model receives empty string as input and is trained only on the labels
    \item For each instance in the dataset, calculate PVI as the difference in the negative log probabilities of the correct answer assigned by the two models
\end{enumerate}

The original implementation is only for text classification, so we adapt it for Extractive QA. In extractive QA, the ``label'' is not one class from a closed set, but rather a sequence of tokens in the input. Therefore we can't feed empty input to the model. To simulate ``null'' input, we feed only the context, but withhold the question. We calculate the probability of the answer in the same way as with ambiguity.

\paragraph{Discriminability}
For SQUAD we don't calculate discriminability ourselves. We instead use the implementation from \citet{rodriguez-etal-2021-evaluation} available at \url{https://www.pedro.ai/leaderboard-acl2021}. 
For MNLI, we use Py-IRT \cite{Lalor_2023} to calculate the discriminability of the data using 10 different models, a standard 2pl model configuration and train the IRT model for 100 epochs. 

\paragraph{Length} 
For SQUAD, we calculate the length of the context as a number of tokens. For MNLI, we calculate the sum of the lengths of the premise and the hypothesis.

\paragraph{Label Noise}
We define ``label noise'' as inverse inter-annotator agreement. We calculate the annotator agreement (in [0,1] range) and then obtain noise as (1 - agreement). Label noise of 0 corresponds to 100\% agreement, while label noise of 1 corresponds to 0\% agreement.

For SQUAD, we calculate the pairwise agreement between any 2 annotators in terms of F1 token overlap. We then aggregate across all pairs to obtain annotator agreement for the pair. This approach is inspired by the way models are evaluated in F1 setting.
For MNLI, we calculate the agreement as the number of annotators that select the majority label.

We test two different ways of obtaining the noise feature: in the ``simple'' setting we just take the inverse agreement as it is. In the ``machine learning'' setting, we train a distilbert-base model to predict ``inverse agreement'' from the text input and we use the prediction from the model.

For SQUAD, we experimented with both configurations, as the way we calculate agreement gives a continuous distribution of noise. The results reported in the paper for SQUAD are using the ``simple'' setting. For MNLI, the ``simple'' setting give three discrete values (0.6, 0.8, and 1), which are difficult to use directly. The results in the paper for MNLI are using the ``machine learning'' setting.

\paragraph{Perplexity}
We calculate perplexity using a pretrained GPT2-large model. Calculating perplexity on a single text is a straightforward task. Calculating perplexity on a task that involves pairs of text (like Extractive QA or NLI) is non-trivial and to the best of our knowledge has not been defined before.

We considered three variants of calculating the perplexity: 1) we can calculate the perplexity of the two text concatenated together; 2) we can calculate the perplexity of only one of the text; 3) we can calculate ``conditional'' by measuring the likelihood of the question (or the hypothesis) given the context (or the premise). We chose to implement the third option, as we believe it makes the most sense in the context of the tasks and the datasets.

\paragraph{Feature Scaling and Outliers}
For easier comparison and visualization, we scale all features to [0--1] range, using MinMax linear scaler. We clip the top and bottom 2\% of the values to reduce the impact of outliers to the scaled distributions.

\paragraph{Code Implementation and Data}
All scripts for feature extraction, all stratified and random data splits, and all experimental analysis and results will be made available at \url{https://github.com/venelink/benchmark_transparency}.

\paragraph{Computational Resources}
The data features were calculated using a single Nvidia V100 or A100 GPU. The total GPU time for all features for both datasets was less than 48 hours.

\section{Stratified Sampling and Bootstrapping}
\label{app:boot}

\paragraph{Stratified Sampling}
To obtain the data samples for each data dimension, we:

\begin{enumerate}
    \item Obtain the values corresponding to 10th, 20th ... 90th percentile
    \item Take all instances with feature value between [0-10p]; [10p-20p]... [90p-1]
\end{enumerate}

Note that we take 10 datasets of equal size rather than taking datasets that correspond to 10\% of the scores (i.e., 0.1 in the scaled vesion of the features). We decided to take percentile-based approach rather than value-based approach due to the skewed distribution of values.

For ``data noise'' in SQUAD, approximately 50\% of the instances had value of 0. To avoid having 5 datasets with the same data distribution, we put all 0-noise instances in one data sample and distributed the remaining 50\% in 9 smaller datasets.

\paragraph{Bootstrapping and Statistical Significance (F1)}

\begin{algorithm}[h]
\caption{Calculating statistical significance of F1 variance for feature-based data samples}\label{alg:f1}
\begin{algorithmic}

\Procedure{BootstrapF1}{$\textbf{M},\textbf{D}$}

    \State \Comment{\textit{input: Model \textbf{M}; Dataset \textbf{D} of size \textbf{n}}}

    \For{trial $t_i$; i $\in$ [0, 1 $\ldots$ 199] }

        \State $D_i$ = RandomSample(\textbf{D},size=\textbf{n}$\div$10)
    
        \State $F1_i$ = Evaluate(\textbf{M},$D_i$)
    \EndFor
    \State $F1_{all}$ = [$F1_0$, $F1_1$ $\ldots$ $F1_{199}$]
    \State \Comment{\textit{Random variance of F1 for \textbf{M}}}
    \State $LB_M$ = ScoreAtPerc($F1_{all}$,2.5)
    \State $UB_M$ = ScoreAtPerc($F1_{all}$,97.5)
\EndProcedure
\\
\Procedure{$F1_{Feat}$}{$\textbf{M},\textbf{D},LB_M,UB_M,c$}

    \State \Comment{c - data dimension (e.g., ``Difficulty'')}
    \State [$D_{c0}$, $D_{c1}$ $\ldots$ $D_{c9}$] = FeatSample(\textbf{D},c)
    \For{$D_i$ in [$D_{c0}$, $D_{c1}$ $\ldots$ $D_{c9}$]}
        \State $F1_i$ = Evaluate(\textbf{M},$D_i$)
        \If{$F1_i$ < $LB_M$ OR $F1_i$ > $UB_M$}
            \State significant $\gets$ significant + 1
        \EndIf
    significant $\gets$ significant $\div$ 10
    \EndFor
    
\EndProcedure

\end{algorithmic}
\end{algorithm}

We used bootstrapping to determine whether the observed variance in F1 w.r.t. data distribution is statistically significant. Bootstrapping is non-parametric and avoids any assumptions about the data distribution.
Algorithm \ref{alg:f1} demonstrates the process for a single model \textbf{M}.

First, we determine the ``expected random variance'' in \textsc{BootstrapF1}. We randomly sample 200 test sets from the dataset \textbf{D}, each with size 10\% of \textbf{D}. We calculate the F1 score of \textbf{M} on all 200 random sets. To obtain the two-tailed statistical significance w.r.t. the ``expected random variance'' we calculate the values at 2.5 and 97.5 percentiles. Any F1 score outside of the range [val(2.5) : val(97.5)] is not generated by a random sampling with a probability \textit{p} < 0.05.

\textsc{$F1_{Feat}$} calculates the statistical significance of F1 variance for a model \textbf{M} and a data dimension \textbf{c}. First, we use stratified sampling to obtain 10 dataset with increasing intensity of \textbf{c}. Then, we calculate the F1 score of \textbf{M} on each of the 10 datasets and compare the values to the ``expected random variance''. We count the number of values (out of 10) that are significantly different from random. This indicates how sensitive is the model \textbf{M} to changes in the distribution w.r.t. \textbf{c}. We also measure the range of F1 (difference between best and worst performance across the 10 test sets) and the standard deviation ($\sigma$) of F1 across the 10 test sets for additional perspective on model consistency.

We repeat the process for all models, using the same 200 random and 10 feature datasets and aggregate the significance scores to obtain the effect that each data dimension has on the F1 score of a model.

\begin{algorithm}[h]
\caption{Calculating statistical significance of rank variance for feature-based data samples}\label{alg:tau}
\begin{algorithmic}

\Procedure{BootstrapRank}{$\textbf{$M_{all}$},\textbf{D}$}

    \State \Comment{\textit \textbf{$M_{all}$ = [\textbf{$M_0$}, \textbf{$M_1$} $\ldots$ \textbf{$M_{124}$}]}}

    \For{trial $t_i$; i $\in$ [0, 2 $\ldots$ 199] }

        \State $D_i$ = RandomSample(\textbf{D},size=\textbf{n}$\div$10)
    
        \State [$R_0$, $R_1$ $\ldots$ $R_{124}$] = EvalRank(\textbf{$M_{all}$},$D_i$)
        \State $KT_i$ = KTau([$R_0$, $R_1$ $\ldots$ $R_{124}$],$R_{ref}$)
        \State \Comment{$R_{ref}$ = mean rank from bootstrap}
    \EndFor
    \State $KT_{all}$ = [$KT_0$, $KT_1$ $\ldots$ $KT_{199}$]
    \State \Comment{\textit{Random variance of ranking}}
    \State $LB_{kt}$ = ScoreAtPerc($KT_{all}$,2.5)
    \State $UB_{kt}$ = ScoreAtPerc($KT_{all}$,97.5)
\EndProcedure
\\
\Procedure{$Rank_{Feat}$}{$\textbf{$M_{all}$},\textbf{D},LB_{kt},UB_{kt},c$}

    \State [$D_{c0}$, $D_{c1}$ $\ldots$ $D_{c9}$] = FeatSample(\textbf{D},c)
    \For{$D_i$ in [$D_{c0}$, $D_{c1}$ $\ldots$ $D_{c9}$]}
        \State [$R_0$, $R_1$ $\ldots$ $R_{124}$] = EvalRank(\textbf{$M_{all}$},$D_i$)
        \State $KT_i$ = KTau([$R_0$, $R_1$ $\ldots$ $R_{124}$],$R_{ref}$)
        \If{$KT_i$ < $LB_{kt}$ OR $KT_i$ > $UB_{kt}$}
            \State significant $\gets$ significant + 1
        \EndIf
    \EndFor
    
\EndProcedure

\end{algorithmic}
\end{algorithm}

\paragraph{Bootstrapping and Stat. Significance (Rank)}
We use bootstrapping to quantify the significance of rank variance by looking at the consistency of the ranking of all systems. The process is described in Algorithm \ref{alg:tau}. 

First, we determine ``expected random variance of ranking'' in \textsc{BootstrapRank}. For each of the 200 test sets we calculate the relative ranking of all models and then compute the ``rank distance'' w.r.t. the reference ranking $R_{ref}$ using Kendall's Tau. The reference ranking $R_{ref}$ is the mean rank of each system across all random samples. We calculate the 2.5 and 97.5 percentile of the 200 Kendall's $\tau$ scores. Any ranking with a $\tau$ outside of [val(2.5) : val(97.5)] is not generated by a random sampling with a probability \textit{p} < 0.05. 

\textsc{$Rank_{Feat}$} calculates the statistical significance of ranking variance for a data dimension \textbf{c}. We use stratified sampling to obtain 10 disjointed datasets with increasing intensity of \textbf{c}, calculate the model ranking and compute the $\tau$ w.r.t. $R_{ref}$. We count the number of datasets where the $\tau$ is significantly different from the random.

\section{Models Used in Evaluation}
\label{app:models}
For the experiments with SQUAD, we use publicly available instance-level predictions from \url{https://rajpurkar.github.io/SQuAD-explorer/}. We download all model predictions and filter out models with F1 above 92 or below 77 to obtain a set of 125 models. The filtering is for visualization purposes and for reducing the impact of outliers. We run the statistical significance tests on all models to ensure that the filtering does not impact the reported results and the conclusions that we draw.

For MNLI, we used publicly available pretrained and finetuned models with different architectures, as available on the huggingface model repository. The list of models that we used is as follows:

\begin{itemize}
    \item Albert (TehranNLP/albert-base-v2-mnli)
    \item Bart-Large (facebook/bart-large-mnli)
    \item Bert-Base (TehranNLP/bert-base-cased-mnli)
    \item Deberta (MoritzLaurer/DeBERTa-v3-large-mnli-fever-anli-ling-wanli)
    \item Distilbert (SEISHIN/distilbert-base-uncased-finetuned-mnli)
    \item Distilroberta (boychaboy/MNLI\_distilroberta-base)
    \item Electra (TehranNLP/electra-base-mnli)
    \item Roberta-Base (TehranNLP-org/roberta-base-mnli-2e-5-42)
    \item Roberta-Large (roberta-large-mnli)
    \item Xlnet (TehranNLP/xlnet-base-cased-mnli)
\end{itemize}

We use the models to score the MNLI-val-matched set without further finetuning or modifications. We used a Nvidia v100 GPU and the process of inference took approximately 1 hour. 

\end{document}